\documentclass[journal]{IEEEtran}

\usepackage{amsmath,amssymb,amsfonts}
\usepackage{algorithmic}
\usepackage{algorithm}
\usepackage{array}
\usepackage[caption=false,font=normalsize,labelfont=sf,textfont=sf]{subfig}
\usepackage{textcomp}
\usepackage{stfloats}
\usepackage{url}
\usepackage{verbatim}
\usepackage{graphicx}

\usepackage{enumitem}

\usepackage[T1]{fontenc}
\usepackage[utf8]{inputenc}
\usepackage[english]{babel}
\usepackage[usenames,dvipsnames,table]{xcolor}
\usepackage[babel]{csquotes}
\usepackage{relsize}
\usepackage{comment}
\usepackage{bm}
\usepackage{rotating}
\usepackage{makecell}
\usepackage{cite}

\newcommand{\redcite}[1]{[\textcolor{red}{\citenum{#1}}]}

\usepackage{tikz}
\usepackage{siunitx}
\usepackage{braket}
\usepackage{pgf}
\usepackage{changepage}
\usepackage{booktabs}
\usepackage{multirow}
\usepackage{bigdelim}
\usepackage{threeparttable}
\usepackage{wrapfig}

\usepackage{listings}
\usepackage{etoolbox}
\usepackage{pgfplotstable}
\usepackage{import}
\usepackage[symbol]{footmisc}
\usepackage{lineno}

\usepackage[hidelinks]{hyperref}

\usepackage{tabularx}

\usepackage[nostamp]{draftwatermark}
\SetWatermarkText{\textsc{Confidential}}
\SetWatermarkScale{1}

\usetikzlibrary{arrows.meta,%
decorations.pathreplacing,%
positioning,plotmarks,%
fit,shapes.symbols,%
shapes.arrows,shapes.geometric,%
shapes.misc,backgrounds,%
graphs,arrows.meta,shadows}
\pgfplotsset{compat=1.18}

\newcommand{\myparagraph}[1]
{\textbf{#1}\quad}

\renewcommand{\arraystretch}{1.6}

\newcounter{guideno}
\renewcommand{\theguideno}{\textbf{G\arabic{guideno}}}
\DeclareRobustCommand{\guideline}[1]{%
   \refstepcounter{guideno}%
   \smallskip\noindent\textbf{\theguideno\ -- #1}\smallskip\par}

\newcounter{futureno}
\renewcommand{\thefutureno}{\arabic{futureno}}
\DeclareRobustCommand{\futuredir}[1]{%
   \refstepcounter{futureno}%
   \smallskip\noindent\textbf{\thefutureno) #1}\smallskip\par}

\newcommand{\dof}{\textsc{Dof}}
\newcommand{\modulename}{shaper}
\newcommand{\modulenames}{shapers}

\newcommand{\Modulename}{Shaper}
\newcommand{\Modulenames}{Shapers}

\newcommand{\ModuleName}{Shaper}

\newcommand{\mathsub}[1]{_\mathit{#1}}

\newcolumntype{M}[1]{>{\centering $}m{#1}<{$}}
\newcolumntype{P}[1]{>{\centering $}p{#1}<{$}}
\newcolumntype{L}[1]{>{\raggedright\arraybackslash}p{#1}}
\newcolumntype{C}[1]{>{\centering\arraybackslash}p{#1}}

\definecolor{lli}{RGB}{255,51,51}
\definecolor{hli}{RGB}{0,128,255}

\def\matlabdir{images/matlab/}
\graphicspath{{images/},{graphs/},{\matlabdir}}

\begin{document}

\title{Myoelectric Control for Upper-Limb Exoskeletons: A Decoder-\ModuleName-Controller Framework, Design Guidelines, and Future Directions}

\author{Davide Costanzi$^{*,1}$, Eldison Dimo$^{*,1}$, Emanuele Feola$^{2}$, Mario Sedilesu$^{1}$, Francesco Pascucci$^{1}$ and Andrea Calanca$^{1}$ 
\thanks{$^{1}$Davide Costanzi, Eldison Dimo, Mario Sedilesu, Francesco Pascucci and Andrea Calanca are with the Department of Engineering for Innovation Medicine, University of Verona, Strada Le Grazie 15, 37134 Verona, Italy (e-mail: davide.costanzi@univr.it, eldison.dimo@univr.it, mario.sedilesu@univr.it, francesco.pascucci\_01@univr.it, andrea.calanca@univr.it).}
\thanks{$^{2}$Emanuele Feola is with the Department of Biomechanical Engineering, University of Twente, 7522 NB Enschede, The Netherlands (e-mail: e.feola@utwente.nl).}
\thanks{$^{*}$These authors contributed equally to this work.
}}

\maketitle

\begin{abstract}
Myoelectric Control Systems (MCSs) enable voluntary control of wearable robots through muscle activity. However, intention decoding, assistance shaping, and low-level actuator control are often not explicitly distinguished, making their contributions difficult to evaluate and compare. This review introduces a functional decoder–shaper–controller framework for continuous myoelectric control of upper-limb exoskeletons. The decoder estimates an intention-related variable from surface electromyography (sEMG), the shaper maps it to a control reference according to an assistance policy, and the controller realizes that reference through actuator control. This decomposition separates intention estimation from assistance-policy design and enables comparison of heterogeneous architectures. Representative approaches are critically examined in terms of module design, calibration, and evaluation. The analysis highlights that the shaper, despite determining the magnitude and shape of assistance, lacks established design and tuning methodologies and is often embedded within the decoder or controller. Key open issues include inconsistent reporting of assistance levels, limited use of interaction-aware stability and passivity tools, and insufficient validation across degrees of freedom, users, and tasks. Building on these findings, this review provides design guidelines for systematic module-wise tuning and identifies future research priorities to bridge the gap between accurate intention decoding and safe, effective, and generalizable assistance under real-world operating conditions.
\end{abstract}


\renewcommand{\thefootnote}{\arabic{footnote}}


\raggedbottom

\section{Introduction}\label{sec:intro}

The growing demand for physical assistance and functional augmentation is driving the development of wearable robotic systems for industrial, rehabilitation, and daily-living applications. In industrial settings, wearable robots can reduce musculoskeletal strain  during repetitive or load-bearing tasks \cite{deLooze2016:ind_exo_review}. In rehabilitation and activities of daily living (ADLs), they can compensate for declining physical capacity in an aging population and support functional autonomy \cite{WHO2021:report}. 

Upper-limb exoskeletons pose distinct challenges compared with their lower-limb counterparts, because of the kinematic complexity of the human arm, which involves multiple degrees of freedom (\dof s), a wide workspace, and highly variable movements requiring dexterity, precision, and adaptability. Effective upper-limb assistance therefore requires control strategies that can continuously and appropriately respond to the wearer's movement intention \cite{Gull2020_review, Bardi2022}. Voluntary control is therefore a key capability for wearable robots. In the literature, it is generally based on measurements of signals originating from the central nervous system, as in brain–machine interfaces, or from the peripheral nervous system, such as those associated with voluntary muscle activation. Among non-invasive approaches, muscle activity measurements are particularly attractive for continuous voluntary control because they provide a direct correlate of voluntary muscle control. Muscle activity can be assessed using several non-invasive sensing modalities, including surface electromyography (sEMG), mechanical, ultrasound-based, optical, and electrical impedance techniques \cite{fu2022:mcs_review,xiao2019:fmg_review,correa-vignais2023:mmg,guo2011:smg-semg_comparison,guo2017:emg_nirs_fusion,zhang2015tomo}. Although the framework we propose is applicable in principle to other sources of peripheral muscle-related information, in the present review we decided to focus on sEMG, the most established and widely used non-invasive technique for continuous voluntary control.

In this paper, a myoelectric control system (MCS) is described as the complete control chain from myoelectric signal acquisition to actuator command execution. An MCS forms an artificial sensorimotor pathway that acts in parallel with the biological limb, as in exoskeletons, or functionally replaces it, as in prostheses. MCSs are a mature technology in prosthetic applications, with commercial systems enabling control of multi-\dof\ artificial limbs \cite{geethanjali2016:myoelectric_hands, Iqbal2018}.
In this context, wearers can adapt to and successfully use non-physiological mappings between available myoelectric signals and prosthetic motor commands, and can develop new motor representations through training \cite{ison2014:learning_control}.

Applying MCSs to assistive exoskeletons introduces additional constraints. Unlike prostheses, which replace missing limbs and act in series with the body, exoskeletons operate in parallel with the biological limb. They therefore require appropriate mechanical alignment and compliance, as well as a high degree of temporal and functional coherence between the user's muscle activity, movement, and the assistance provided by the device, to avoid disrupting natural or residual movements \cite{massardi2023hazards}. Excessive delay or inappropriate device responses may also interfere with the wearer's sense of agency, or perception of having control over the device \cite{wen2015:action_delay}.

\begin{table*}[htp]
\centering
\small
\setlength{\tabcolsep}{3.0pt}
\renewcommand{\arraystretch}{1.03}

\rowcolors{3}{gray!10}{white}

\caption{Representative recent reviews on topics related to continuous myoelectric control.}

\begin{tabularx}{\linewidth}{
    p{0.95cm}
    p{0.8cm}
    p{3.7cm}
    X
}
\toprule
\rowcolor{white}
\textbf{Ref.} &
\textbf{Year} &
\textbf{Architecture} &
\textbf{Topics} \\
\midrule

\redcite{hakonen2015:myoelectric_review} & 2015 & Continuous/discrete decoder &
Recommended practices and future research on myoelectric interfaces\\

\redcite{novak2015:sensors} & 2015 & Decoder, controller &
Sensor fusion methods for wearable robots\\


\redcite{li2020:review} & 2020 & Decoder, controller &
Key technologies for human-robot interaction based on sEMG \\

\redcite{Bao2022} & 2022 & Decoder &
Machine-learning and deep-learning methods for robust, adaptive, and reliable upper limb motion estimation for myoelectric control \\

\redcite{fu2022:mcs_review} & 2022 & Decoder, controller &
MCSs for upper limb wearable robotic exoskeletons and exosuits \\

\redcite{lloyd2023} & 2023 & Decoder &
Neuromusculoskeletal biomechanics: neural control, muscle--tendon dynamics, musculoskeletal modeling, and digital-twin applications \\

\redcite{wei2024:emg_review} & 2024 & Decoder &
Model-based and model-free approaches for continuous upper limb motion intention prediction from sEMG \\

\redcite{xiong2024:intuitive_hrei_emg} & 2024 & Continuous/discrete decoder &
Intuitive Human-Robot-Environment Interaction (HREI) with sEMG \\


\redcite{toroossaba2025myoelectric} & 2025 & Decoder, controller &
MCSs for rehabilitative and assistive soft exoskeletons, with emphasis on soft-robotic integration and real-time adaptation \\

\bottomrule
\end{tabularx}
\label{tab:mcs_reviews}
\end{table*}

Recent reviews have provided valuable overviews of EMG-driven control for upper-limb exoskeletons and exosuits, including classifications of MCSs based on high-level control modalities \cite{fu2022:mcs_review, toroossaba2025myoelectric}, or on signal acquisition, preprocessing, and machine-learning-based decoding methods \cite{Sul2025}.
However, existing perspectives generally treat the control pipeline as a unified process and do not explicitly separate intention decoding, assistance shaping, and actuator control. As a result, functionally distinct roles may be grouped together, making architectures difficult to compare and hindering systematic design. Table~\ref{tab:mcs_reviews} summarizes the relevant reviews considered in this work, reporting their architectural conceptualization and main topics.

From our critical analysis of MCSs for upper-limb exoskeletons, we propose a three-module functional taxonomy, separating the control architecture into a \textit{decoder}, which estimates movement intention from \mbox{sEMG}; a \textit{\modulename}, which transforms this estimate into an appropriate assistance command or reference; and a \textit{controller}\footnote{The term ``controller'' is used here exclusively to denote the low-level control module within the MCS.}, which converts this command into actuator actions. Decoder, \modulename, and controller address fundamentally different design problems: intention estimation, human-in-the-loop assistance shaping, and low-level control, respectively. 
This functional separation allows control architectures to be compared independently at the levels of intention estimation, assistance generation, and actuator execution, regardless of the specific algorithms or sensing technologies used to implement them.

This review is structured as a critical synthesis rather than a systematic or historical survey. Instead of seeking exhaustive coverage, we analyze representative contributions that illuminate recurring architectural choices and functional roles within the proposed taxonomy. The objective is to clarify the functional relationships among the roles of the three modules, identify limitations and open challenges, and derive design guidelines and future research directions.

The remainder of the paper is organized as follows. The rest of this section describes the literature search strategy. Section \ref{sec:taxonomy} presents the proposed taxonomy and its motivations. Sections \ref{sec:decoders}, \ref{sec:adapters}, and \ref{sec:controllers} examine the three modules in detail. Section \ref{sec:examples} discusses representative examples from the literature within the proposed framework. Section \ref{sec:guidelines} provides design guidelines, while Section \ref{sec:future_directions} outlines possible future research directions.

\subsection{Literature Search Methodology}\label{sec:intro:method}

We conducted a literature search across major electronic databases, including Scopus, PubMed, IEEE Xplore, and Google Scholar, focusing on MCSs for upper-limb exoskeletons/exosuits implementing intention-based continuous/proportional control using sEMG as the primary input source. Consequently, we excluded lower-limb and back devices, passive orthoses, and active devices that did not include an intention-based continuous/proportional control component. Rehabilitation, assistive, and industrial devices were considered.

The reviewed literature covers primarily the period 2015-2026, but previous seminal studies were included when they provided influential methodological or conceptual contributions. The literature search was last updated on August 31, 2026.
Representative search terms included combinations of \textit{myoelectric}, \textit{electromyography}, \textit{EMG}, \textit{proportional control}, \textit{continuous control}, \textit{exoskeleton}, \textit{orthosis}, \textit{exosuit}, \textit{upper limb}, \textit{arm}, \textit{shoulder}, \textit{elbow}, \textit{wrist}, and \textit{hand}.
After identifying the three functional modules, we performed additional focused searches targeting decoders, \modulenames, and controllers to identify representative studies that address each functional role. 

\begin{figure*}
    \centering
    \includegraphics[width=2\columnwidth]{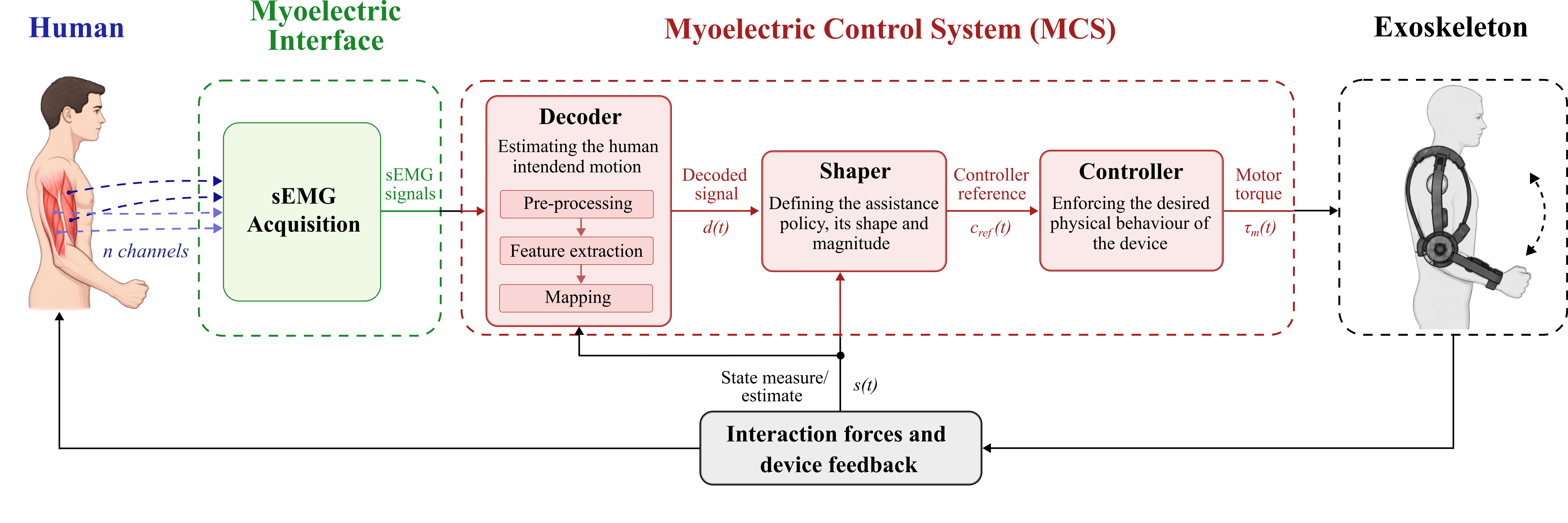}
    \caption{Overview of the proposed three-module architecture for myoelectric control of upper-limb exoskeletons, comprising the decoder, shaper, and controller, together with the myoelectric interface and human–exoskeleton feedback pathways.}
    \label{fig:taxonomy:iface}
\end{figure*}

\subsubsection{Inclusion Criteria}

Studies were included if their control architecture could be clearly identified and classified within the proposed taxonomy, or if they provided relevant architectural insights. Preference was given to studies reporting experimental validation with human participants or, where appropriate, simulation results supported by a clear theoretical contribution. Peer-reviewed journal articles and papers published in established conferences in robotics, biomedical engineering, and rehabilitation engineering were prioritized. Studies lacking sufficient methodological detail, a clear description of the control pipeline, or adequate validation were generally excluded.

\section{A Three-Module Taxonomy for Myoelectric Control Systems}\label{sec:taxonomy}

\subsection{The Functional Decomposition}\label{sec:taxonomy:arch}

The proposed taxonomy decomposes MCS architectures into three functional modules as represented in Figure \ref{fig:taxonomy:iface}. These modules are often implicitly present as functional roles, even when the MCS does not explicitly implement them as distinct components.

The \emph{decoder} has the role of estimating an intention-related control variable from myoelectric activity. In general, multiple sEMG signals recorded at different locations are processed to estimate a voluntary control signal $d(t)$ representing the user's intended movement or effort. For continuous proportional control, decoders are often calibrated to approximate a measurable target quantity $d^*(t)$, providing the decoded signal $d(t)$ with a defined physical or behavioral interpretation. Typical targets include force or torque associated with muscular activation, such as isometric torque \cite{Camardella2021,dimo2026:partial_agc,costanzi2023:personal_myo} or a dynamic torque profile obtained through inverse dynamics \cite{quesada2026:emg_torque}. Because the mapping between muscle activation and forces depends on biomechanical configuration, decoders may also incorporate kinematic signals $s(t)$ from accelerometers, IMUs, encoders, or other sensors \cite{dwivedi2022:emg_fusion_review}. The decoder can be further decomposed into: (1) preprocessing of raw sEMG signals, (2) feature extraction, and (3) mapping of the resulting features to $d(t)$.

The \emph{\modulename} transforms the decoded signal $d(t)$ into a control reference $c\mathsub{ref}(t)$. Its role is to determine how the estimation of human intention is translated into robotic assistance, including both the \emph{shape} and the \emph{magnitude} of the assistance response. It may be implemented as a simple gain or as a more complex mapping, including admittance models \cite{lobo-prat2016:emg_impl} and linear or non-linear dynamics \cite{costanzi2023:personal_myo,TreussartVignais2020,Dimo2026MyographyGravityCompensation}.
Importantly, $d(t)$ and $c\mathsub{ref}(t)$ do not need to represent the same physical quantity. For example, a decoder may estimate muscular torque, whereas the \modulename\ may generate a motor-velocity reference. Thus, the \modulename\ provides the functional interface between intention estimation and robotic assistance.

The \textit{controller} regulates a physical quantity $c(t)$ toward the reference $c_{\mathrm{ref}}(t)$ generated by the shaper, producing the corresponding actuator commands, such as motor torque $\tau_m(t)$. Typical implementations include position, velocity, force, impedance, and admittance control, with design objectives such as tracking accuracy, stability, and appropriate interaction dynamics.



For simplicity, the formulation above considers a single-\dof\ system, but it readily extends to multi-\dof\ and MIMO architectures. Multiple single-joint MCSs may operate in parallel, or the three functional roles may be implemented across multiple joints within a unified architecture \cite{sedighi2025meta,Sedighi2024,oghogho2022deep}.

\begin{figure*}[tp]
    \centering
    \includegraphics[width=2\columnwidth,trim=0mm 0mm 8mm 0mm,clip]
    {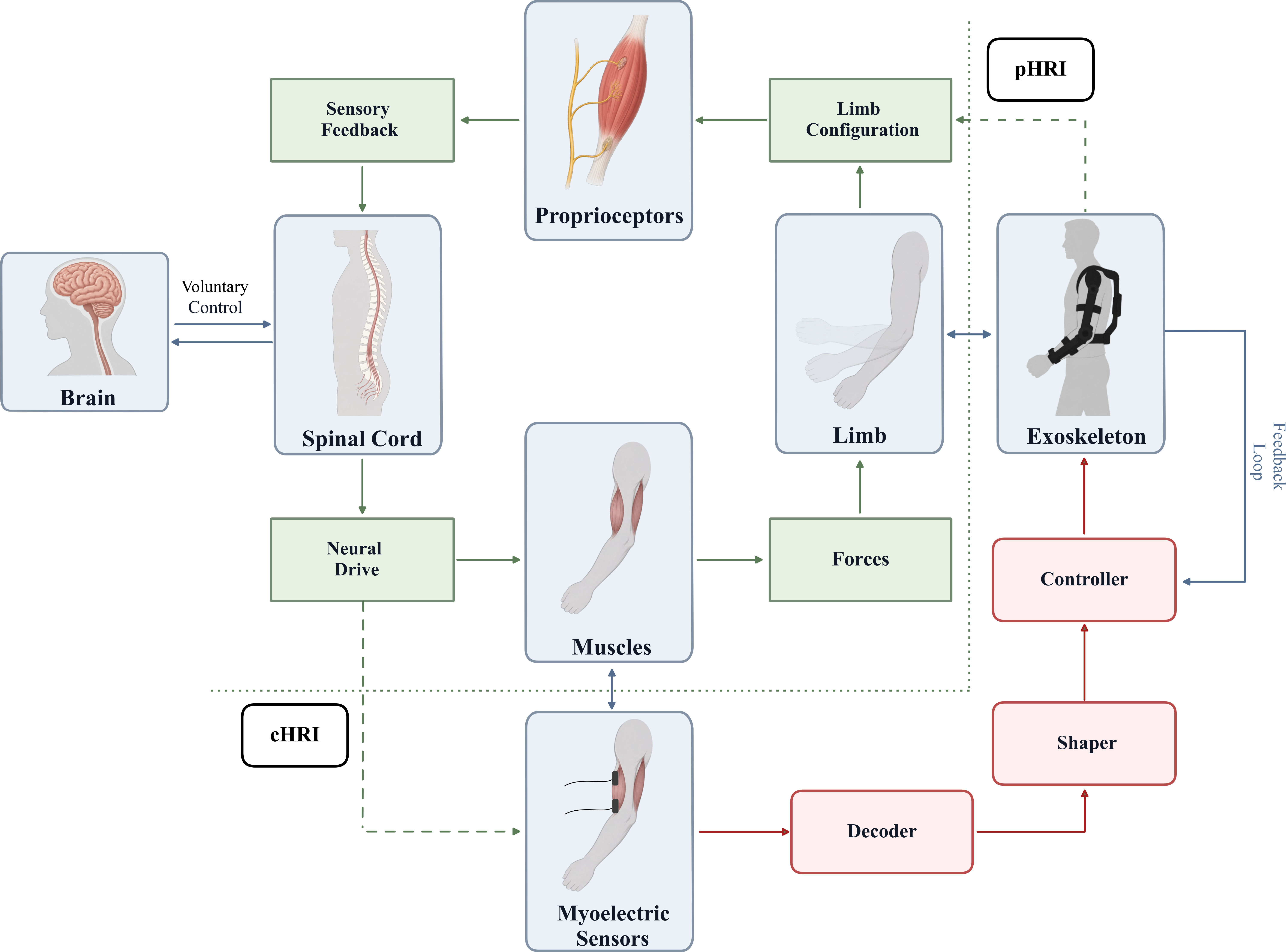}
    \caption{Human motor control loop for a subject wearing an exoskeleton. Blue boxes denote physical components, green boxes indicate information transmitted between components, and red boxes represent roles in the MCS taxonomy.}
    \label{fig:taxonomy:human_ctrl_exo}
\end{figure*}

\subsection{Motivations for Three Separate Modules}\label{sec:taxonomy:motivation}

The three modules have distinct functional roles and address different design problems: estimation of user intent, generation of the desired assistance response, and closed-loop regulation of the robotic system. 
These design problems can be treated separately at the architectural level, although their behavior remains coupled through the human–robot interaction loop.

For the decoder, calibration against a common measurable target $d^*(t)$ provides a basis for separating decoding performance from downstream assistance scaling. Consider a simple example involving two decoders producing $d_1(t)$ and $d_2(t) = 10 d_1(t)$. Assuming the same temporal structure, their ability to represent the intended movement is qualitatively equivalent. However, the second output would produce a larger assistance response if passed directly to the same downstream mapping, so they may have different system-wise performance. This difference should be more appropriately attributed to assistance scaling rather than to decoding capability, because the outputs of the two decoders have identical shape and only differ by scale. Calibrating different decoders against the same target $d^*(t)$ provides a principled basis for exchanging and comparing them independently of the assistance scaling. This distinction is particularly relevant for continuous proportional MCSs, in which decoder output amplitude directly affects the commanded assistance. Some studies do not explicitly calibrate the decoder to a target quantity
\cite{Peng2024:myomo,oghogho2022deep,Lambelet2017_eWrist,lobo-prat2016:emg_impl,lenzi2012:emg}.

The controller regulates a physical variable $c(t)$ towards $c\mathsub{ref}(t)$ by generating the actuator command $\tau_m(t)$, according to established control objectives. Its tuning therefore does not depend on the internal mapping used by the decoder. In particular, the controller performance can be evaluated independently of the decoder's fitting procedure, even though the resulting closed-loop behavior is ultimately coupled with the human and the upstream modules.


The \modulename\ establishes the relationship between the decoded intention and the controller reference, thereby determining how the user's effort is translated into robotic assistance. Unlike decoder calibration and controller design, assistance shaping lacks established methodologies for selecting and tuning the desired assistance response, which is often determined empirically or through manual adjustments \cite{fu2022:mcs_review, missiroli2020:assistance_level}. Treating the \modulename\ as a distinct functional role therefore makes the assistance policy explicit and allows its effects to be examined separately from decoding accuracy and low-level control performance.

In some implementations, the \modulename\ may reduce to a simple gain, potentially embedded within the controller. Nevertheless, keeping it as a logically separate module remains conceptually important, as its action has a dominant effect on human effort, fatigue and comfort \cite{dimo2026:partial_agc}. Unlike the controller, which is designed only based on exoskeleton dynamics, and the decoder, which is designed only based on the relationship that maps myoelectric signals to a measurable quantity $d^*(t)$, the \modulename\ directly interferes with the user’s sensorimotor control system and significantly influences the overall human-robot behavior. Inappropriate tuning may compromise usability: as reported in \cite{Kwon2014}, excessive amplification of the decoded signal can hinder the user’s ability to operate the device effectively.




\subsection{Relationship with the Human Sensorimotor Control Loop}\label{sec:taxonomy:sensorimotor}

An upper-limb exoskeleton controlled by an MCS operates in parallel with the user’s biological motor control system. The resulting human-robot interaction (HRI) occurs at both the physical level (pHRI), through mechanical interaction, and the cognitive level (cHRI), through voluntary control \cite{pons2008:wearable}.

Figure \ref{fig:taxonomy:human_ctrl_exo} illustrates the resulting sensorimotor loop. The central nervous system (CNS) generates motor commands through higher-level voluntary control and spinal sensorimotor processes. These commands propagate through the peripheral nervous system (PNS), resulting in muscle activation and movement.  The resulting sEMG is measured by the myoelectric interface and processed by the MCS to generate assistive commands. The exoskeleton therefore creates an artificial sensorimotor pathway in which myoelectric activity associated with voluntary effort is translated into mechanical assistance that modifies the resulting limb motion and, consequently, the sensory feedback available to the user.

The role of this artificial loop depends on the application. In industrial applications, the exoskeleton supports a healthy motor system \cite{moeller2022:industrial_arm_exo_review}. 
In assistive applications, the exoskeleton can compensate for insufficient muscle force to perform a movement \cite{eksobionics2017:ekso,pedrocchi2013:mundus}.
Rehabilitative applications introduce additional challenges because voluntary myoelectric activity may be weak or altered by neurological impairment. Nevertheless, recent evidence shows that voluntarily modulated motor-unit activity can remain detectable using high-density electromyography (HD-EMG) even in some individuals with motor-complete cervical spinal cord injury \cite{oliveira2024direct}. This suggests that myoelectric control may be feasible even in individuals with severe motor impairments, although its practical implementation requires further investigation.

Assistive and rehabilitative applications are not mutually exclusive, as the same robotic platform may be used for both purposes. In rehabilitation, sEMG may serve not only as an input for intention-based control but also to assess muscle activation, co-contraction, and coordination, providing information to adapt the rehabilitation task \cite{proietti2016:control_review}.

\section{The Decoder Module}\label{sec:decoders}

Decoders can be distinguished according to whether they generate discrete or continuous outputs. \emph{Discrete} decoders, also referred to as \emph{pattern-recognition-} or \emph{classification-based} decoders, map sEMG patterns into a finite set of actions or control modes \cite{fougner2012:prostheses_terminology}, whereas \emph{continuous} decoders estimate a continuously varying control variable. Because the present review focuses on intention-based proportional control, we primarily consider continuous decoders. Discrete decoders are therefore outside the scope of this section, although they may nevertheless be combined with continuous decoding to select a mode or trigger predefined actions \cite{McDonald2020hx, dallagasperina2021:coop_ctrl_rehab}.

A typical sEMG decoding pipeline comprises preprocessing, feature extraction, and mapping. Recent reviews have already provided comprehensive overviews of sEMG signal preprocessing and feature extraction techniques for myoelectric control \cite{Sul2025, Phinyomark2018, li2020:review}. 
These topics are therefore outside the scope of the present review. In this review, we focus on the mapping between sEMG-derived information and the decoded target $d(t)$, with particular emphasis on decoders' underlying principles, implementation, main challenges and limitations. The approaches considered below represent different modeling paradigms rather than mutually exclusive classes.

Table \ref{tab:decoders:overview} summarizes the continuous decoder categories considered in this section. Their main characteristics and trade-offs are synthesized in Section~\ref{sec:decoders:discussion}.

{\renewcommand{\labelitemi}{$\cdot$}

\begin{table*}[!t]
\begin{adjustwidth}{-15mm}{-15mm}
\centering
\small
\setlength{\tabcolsep}{4pt}
\renewcommand{\arraystretch}{1.25}

\caption{Overview of the continuous decoder paradigms considered in this review. Generalization, calibration burden, and computational burden are qualitative considerations that depend on the prediction target, model complexity, available data, and implementation.}
\label{tab:decoders:overview}

\begin{tabular}{
    @{}
    L{2.40cm}
    @{}
    >{\columncolor{gray!10}}L{2.50cm}
    L{2.50cm}
    >{\columncolor{gray!10}}L{2.50cm}
    L{2.75cm}
    >{\columncolor{gray!10}}L{2.51cm}
    L{2.50cm}
    @{}
}

\toprule

\textbf{Decoder Paradigm} & \textbf{Empirical} & \textbf{Linear Regression} & \textbf{Non-linear Regression} & \textbf{NMS Models} & \textbf{ML Models} & \textbf{Physics-Informed ML Models} \\

\midrule

\textbf{Typical Target} & No explicit target & $d^*(t)$: force, joint torque, kinematics & $d^*(t)$: force, joint torque, kinematics & $d^*(t)$: force, joint torque; kinematics possible & $d^*(t)$: kinematics; force, joint torque possible & Potential $d^*(t)$: force, joint torque, stiffness, muscle states, kinematics \\

\textbf{Physiological Structure} & None & None & Optional & Explicit & 	No explicit physiological structure & Physical constraints through loss functions or model architecture \\


\textbf{Generalization} & Very low & Improved when fitting force/torque rather than kinematics & Improved when fitting force/torque rather than kinematics & 	Potential robustness across payloads; model- and calibration-dependent & Strongly dependent on training data distribution & Potential improvement over conventional ML; not yet demonstrated in complete MCSs \\


\textbf{EMD Handling} & Generally ignored & Generally ignored & Possible through explicit temporal or activation dynamics & Can be explicitly modeled & Implicitly handled through the temporal analysis window/forecasting horizon & Potentially incorporable through physical constraints \\

\textbf{Calibration Burden} & No explicit decoder calibration & Very Low & Low-moderate & Model-dependent & Data- and model-dependent & Data- and model-dependent \\[1mm]

\textbf{Computational Burden} & Very Low & Very Low & Low-moderate & Implementation-dependent & Model-dependent & Model-dependent \\[1mm]

\textbf{Main Advantage / Limitation} & Simplicity vs. limited decoder interpretability & Computational efficiency vs. limited nonlinear modeling capacity & Nonlinear modeling capacity vs. model complexity & Physiological interpretability vs. subject-specific calibration burden & Modeling flexibility vs. training data dependence & Physical constraints vs. optimization complexity \\[1mm]

\bottomrule

\end{tabular}
\end{adjustwidth}
\end{table*}}

\subsection{Empirical Decoders}\label{sec:decoders:empirical}

Empirical approaches directly map the sEMG features to the assistance command $c\mathsub{ref}$, often using manually selected gains \cite{Bos2020:symbihand_dmd, Peng2024:myomo,lenzi2012:emg}, or normalization procedures that constrain the output to a predefined range \cite{lobo-prat2016:emg_impl}. The mapping is therefore not explicitly calibrated against a measurable target $d^*(t)$. Within the proposed taxonomy, such approaches combine decoding and assistance shaping into a single mapping, making it difficult to distinguish decoding performance from the effects of assistance scaling.

\subsection{Linear and Non-Linear Regression Decoders}\label{sec:decoders:linear_nl_regression}

Regression decoders explicitly identify a mapping between sEMG features and a measurable target $d^*(t)$ using calibration data. Multivariate linear regression (MVLR) provides a computationally efficient baseline to estimate quantities such as joint torque \cite{quesada2026:emg_torque}. Position or other kinematic variables can be included as additional predictors to account for configuration dependence \cite{ziai2011:linear_emg_pos}. Unlike linear models, nonlinear regression approaches are better able to capture nonlinear and configuration-dependent EMG–force relationships but at the cost of increased model complexity and calibration requirements \cite{quesada2026:emg_torque,Lee2025,Berning2021}. Some recent nonlinear approaches have also introduced physiological activation dynamics, electromechanical-delay compensation, and state estimation to improve torque or kinematic prediction \cite{wu_neural_network_enhanced_2019,pan2019,lu2019:emg_torque,Lloyd2003,potvin_mechanically_1996}.

\subsection{Neuromusculoskeletal (NMS) Decoders}

Neuromusculoskeletal (NMS) decoders use physiologically informed forward models to map sEMG-derived muscle excitation to biomechanical quantities such as joint torque, kinematics, stiffness, compressive forces, or musculotendon stresses \cite{quesada2026:emg_torque,kawase2017:emg_arm,pan2025enhancedMM,Wang2026:EMG-informed_NMS,arjmand2009comparison,moya-esteban_robust_2022}. Unlike purely data-driven models, they explicitly model activation and contraction dynamics and use musculoskeletal geometry to account for posture-dependent force transmission \cite{lloyd2023,sartori2016:emg_modelling,zajac1989:muscle_model}, providing physiological interpretability and a principled treatment of muscle redundancy \cite{moya-esteban_robust_2022,Hambly2025,Bennett2022}. Subject-specific parameters must be identified through optimization using experimental calibration data, usually with musculoskeletal modeling frameworks such as OpenSim, CEINMS, MOCO, or SCONE \cite{Delp2007,Pizzolato2015,dembia2020:opensim_moco,Geijtenbeek2019}. Calibration procedures typically target joint torques $d^*(t)$ estimated via inverse dynamics \cite{lotti2022:myo_vs_force, feola2023:payload}, or occasionally joint angles \cite{kawase2017:emg_arm}. Model optimization and execution can be computationally demanding, although recent advances in differentiable physics \cite{Hambly2025}, improved optimization \cite{Bailly2021}, and real-time NMS frameworks, such as CEINMS-RT \cite{Sartori2026CEINMSRT}, are substantially reducing their costs.

Comprehensive overviews of NMS modeling are provided in \cite{lloyd2023, bersani2023}.

\subsection{Machine Learning (ML) Decoders}

During the past decade, ML-based decoders have become increasingly prevalent in myoelectric decoding \cite{wei2024:emg_review,fu2022:mcs_review}. Conventional ML approaches are primarily data-driven, learning a direct mapping from sEMG signals to biomechanical variables without predefined physical constraints. 

Various ML techniques have been explored, ranging from classical algorithms \cite{Zhang2025,pan2019}, which rely on manual feature extraction, to deep learning architectures, which can learn representations directly from raw sEMG, utilizing convolutional and recurrent models to capture spatial and temporal dependencies \cite{Sedighi2024,kim2020:emg_angle_torque,ameri_regression_2019,xia_emg_based_2018}, and Reinforcement Learning (RL) approaches \cite{wu2021}.

A major limitation of ML approaches is their dependence on representative training data. Most current ML decoders are subject-specific, requiring extensive individual datasets to generalize effectively \cite{Sul2025}. 
Even minimal changes in electrode placement, task, posture, fatigue, or external load can shift the data distribution and degrade performance \cite{castellini2014:first_pmi_workshop}. Transfer Learning (TL) and Domain Adaptation \cite{Bao2022} have therefore been investigated to reduce the amount of subject- and session-specific data required \cite{wei2024:emg_review,fu2022:mcs_review}.

\subsection{Physics-Informed ML Decoders}

Physics-Informed Machine Learning (PIML) is an emerging approach in myoelectric decoding that combines the predictive power of deep learning with the physiological consistency of NMS models \cite{Meng2025, Ren2025}. Physical knowledge is typically incorporated either through soft constraints in the loss function that penalize violations of biomechanical equations \cite{Taneja2024,Zhang2023distributed,taneja2022,zhang2022}, or through physics-informed architectures that replace selected components of an NMS model with learned functions while retaining analytical components \cite{schulte2022}.

Despite its potential, PIML presents several challenges, including optimization difficulties arising from competing data-driven and physics-based loss terms, as well as potential model mismatch when simplified rigid-body assumptions fail to adequately capture soft-tissue biomechanics \cite{Luo2025, Ren2025}. Moreover, to the best of the authors' knowledge, no studies have yet reported the integration of a PIML decoder into a complete MCS architecture for upper-limb exoskeleton control.

\subsection{Discussion On Decoders}\label{sec:decoders:discussion}
The major difficulty in continuous myoelectric decoding is that the relationship between sEMG and the biomechanical quantity of interest is nonlinear and depends on factors including joint configuration, electrode placement, and muscle fatigue \cite{castellini2014:first_pmi_workshop,ison2014:synergies_review}. The various decoder paradigms therefore represent different compromises between physiological structure, calibration requirements, computational cost, and generalization.

Regression decoders offer useful intermediate trade-offs. Linear regression offers rapid calibration and low computational cost, and can predict physically meaningful targets such as joint torque, but has limited ability to capture nonlinear sEMG-force relationships, particularly in complex multi-DOF joints such as the shoulder \cite{Lavaill2025,quesada2026:emg_torque}. Classic nonlinear regression approaches provide greater flexibility while retaining a relatively simple calibration procedure. Recent results even showed classic nonlinear regression outperforming both linear regression and NMS models in torque-estimation accuracy under single- and multi-joint conditions while maintaining low calibration and computational costs \cite{quesada2026:emg_torque}. 

Physiological models, such as NMS decoders, can capture part of this complexity, but require several subject-specific parameters that are difficult to measure directly \cite{winby2008:nms_scaling,sartori2014:emg_driven_nms,sartori2016:emg_modelling}. Consequently, subject-specific parameter identification can be time-consuming: reported calibration times range from approximately $20$--$60$ minutes for relatively simple models \cite{quesada2026:emg_torque}, to $3.5$ hours \cite{Lavaill2025}, and up to $25.8$ hours in more demanding implementations \cite{Bennett2022}. These values are not directly comparable because calibration protocols, model complexity, and anatomical applications differ across studies. Recent advances in differentiable physics and gradient-based optimization, however, have reduced calibration times by up to $26\times$ compared with a CEINMS-based approach \cite{Hambly2025}.

ML decoders instead avoid explicit physiological modeling by learning the mapping from sEMG to biomechanical variables directly from data. This generally simplifies model formulation, but requires sufficiently large and diverse datasets that span the relevant tasks, loads, and user conditions. Moreover, because biomechanical constraints are not necessarily embedded in the model, ML decoders may produce outputs that are not physically consistent, such as unrealistic joint torques or joint angles \cite{Sedighi2024,sartori2018}. Calibration time is rarely reported explicitly and depends strongly on model complexity, dataset size, and available computational resources, although existing studies suggest that training may require several hours \cite{wu2021,Sedighi2024}.


In terms of generalization, NMS models can exploit their physiological structure to generalize across different payloads and operating conditions \cite{durandau2022,feola2023:payload}, whereas ML models depend more strongly on the training data distribution and may overfit to specific tasks or payloads \cite{wu2021,pan2019,zhang2021}. In principle, this limitation could be mitigated by collecting larger and more diverse datasets. In practice, however, acquiring personalized data across multiple tasks, payloads, fatigue conditions, and electrode placements remains impractical for real-world applications\footnote{sEMG signals are highly sensitive to electrode placement, and even small electrode shifts can require model recalibration \cite{castellini2014:first_pmi_workshop}.}. Pre-training on large multimodal datasets could provide an alternative, but datasets simultaneously covering a broad range of subjects, tasks, electrode configurations, sEMG signals, kinematics, and interaction forces are still largely unavailable. The choice of prediction target can further affect generalization. NMS models typically estimate joint torques, whereas ML decoders often predict kinematic variables such as joint position (see Table \ref{tab:mcs_classification}). Since the same trajectory may require substantially different muscle activation patterns under different payloads, models trained directly on kinematics can become strongly dependent on the task and loading conditions represented in the training data, thereby reducing their ability to generalize to unseen conditions.

PIML approaches may offer another compromise by combining the flexibility of ML with an explicit biomechanical structure. However, they have yet to be applied to complete MCSs, and the additional complexity introduced by physics-based constraints must be justified by measurable improvements in robustness, generalization, or interpretability.

Overall, no single decoder paradigm is universally preferable. Decoder selection should instead be guided by the required output variable, the target user population, the operating conditions, the expected differences between calibration and deployment conditions, the available calibration resources, and computational constraints.

\subsection{Discussion on Electromechanical Delay}\label{sec:decoders:discussion_emd}

Electromechanical delay (EMD) is the delay between muscle stimulation and measurable force generation \cite{cavanagh1979:emd}.
Therefore, sEMG activity precedes force onset, providing an anticipatory time window that can potentially be exploited to partially compensate for the acquisition, computation, communication, and actuation delays of an MCS. EMD is also relevant for ensuring appropriate temporal alignment between user intention and robotic assistance and, ultimately, for preserving the user's sense of agency \cite{wen2015:action_delay}. Despite the importance of EMD in ensuring timely assistance, it is still only marginally considered in existing MCSs.

A single representative physiological value for the EMD is difficult to define. Reported values range approximately from $10$ to \SI{100}{ms} \cite{cavanagh1979:emd}, depending on muscle, fatigue state, joint posture, and muscle-tendon mechanics \cite{hug_muscle_2015,Smith2017,Fagbemigun2023}. The variability is particularly relevant when translating results from healthy populations to pathological conditions, where EMD may be prolonged or asymmetric between limbs \cite{Son2021,Lacourpaille2017}. Moreover, relaxation EMD (R-EMD), defined as the delay between the cessation of sEMG and force decay, can be longer than onset EMD, especially under fatigue \cite{Smith2017-remd}. 
Neglecting R-EMD may cause an interruption of the assistance before the user's force has decayed.

EMD arises from electrochemical and mechanical processes influenced by muscle state and contraction dynamics, making a fixed-delay approximation potentially inaccurate across muscles and operating conditions \cite{Ce2017,Smith2017,Ce2013}. Its dependence on the rate of force development suggests a frequency-dependent phase lag rather than a constant temporal offset \cite{ubeda_electromechanical_2017}.

Different decoder paradigms handle EMD in different ways. NMS models can explicitly include EMD in excitation-to-activation dynamics \cite{savage_neuromusculoskeletal_2023,Wang2026:EMG-informed_NMS}. Standard ML decoders usually represent temporal dependencies implicitly through the analysis window, commonly \SIrange{150}{250}{ms}, which may, however, cause phase mismatch when physiological EMD changes over time \cite{Englehart2003,Farrell2007}. Predictive ML decoders instead use a forecasting horizon to anticipate future force generation and compensate for communication delays and EMD \cite{Koo2025}. Other MCSs do not  explicitly model EMD, but verify that the overall system response time remains below a representative EMD value \cite{lotti2022:myo_vs_force,lotti2020:myocontrol_soft,missiroli2020:assistance_level}.

In some implementations, EMD may remain unmodeled because part of its temporal effect can be masked by delays introduced by signal filtering \cite{McDonald2013:filtering}, activation dynamics \cite{lloyd2023}, or low-level control. However, this compensation is incidental rather than deliberately designed and may not remain effective across different operating conditions.

Finally, EMD should also be considered together with the complete MCS latency: acquisition, preprocessing, decoding, assistance shaping, communication, control, and actuator dynamics jointly determine the temporal relationship between the user's physiological intention and the resulting assistance.

\section{The \ModuleName\ Module}\label{sec:adapters}

The \modulename\ module transforms the decoded movement intention $d(t)$ into a continuous controller reference $c_{ref}(t)$ according to the assistance policy. 
The \emph{assistance policy} defines the magnitude, the temporal behavior, and the physical interpretation of the assistance provided. The \modulename\ therefore determines how decoded intention is transformed into assistance and lies within the artificial sensorimotor loop between the user and the robot (Figure~\ref{fig:taxonomy:human_ctrl_exo}). This mapping can therefore modify not only the magnitude of assistance but also the relationship between the user's effort, the resulting robot motion, and the forces exchanged with the environment.

Unlike decoder calibration and controller design, there are no established guidelines available for selecting or tuning the \modulename\ module.
Existing implementations in the literature span a wide range of approaches, from simple static mappings to more complex dynamic, adaptive and learning transformations.

\subsection{Proportional \Modulenames}

The simplest shaping strategy is proportional mapping:
\begin{equation}
    c_{ref}(t) = \beta d(t),
\end{equation}
where $\beta$ determines the constant scaling between the decoder output and the controller reference.

The first and still commonly used implementation is force-to-force proportional shaping \cite{Rosen2001MyosignalExoskeleton,lenzi2012:emg,lotti2022:myo_vs_force,Li2023_torque_prediction,costanzi2023:personal_myo}. Here, $d(t)$ represents an estimate of the force or torque intended by the user, and the \modulename\ converts it into a force or torque reference for the controller. The exoskeleton therefore acts as a strength amplifier: the robot contribution increases proportionally with the decoded voluntary effort. With this kind of mapping, the gain $\beta$ directly controls the force-augmentation ratio, which is also related to the assistance level, or human-robot force sharing (see Section~\ref{sec:adapters:discussion}).

A less common variant is force-to-velocity proportional shaping \cite{lu2019:emg_torque,Peng2024:myomo}, in which
\begin{equation}
    \dot{q}_{ref}(t) = \beta d(t).
\end{equation}
The decoded effort therefore determines the desired velocity of the assisted joint, while the controller generates the torque required to track that velocity. Compared with force-to-force proportional shaping, the same decoded effort can ideally result in the same velocity under different payloads. The resulting assistive force therefore depends on the coupled dynamics of the user, robot, and task, making the physical interpretation of $\beta$ less direct.

\subsection{Integral \Modulenames}

An integral \modulename\ maps the decoded signal to the rate of change of the controller reference:
\begin{equation}
    \dot{c}_{ref}(t) = \beta d(t).
\end{equation}
With this strategy, the user's effort controls the changes in the reference value. Once the desired value is reached, a constant force, torque, or position can be maintained without requiring sustained muscle activation \cite{costanzi2023:personal_myo}. 

The integral-force variant is particularly suitable for slow and quasi-static anti-gravity assistance \cite{TreussartVignais2020,costanzi2023:personal_myo}.
Integral-position mappings are also possible. When paired with a "stiff" position controller, the user's effort effectively determines the joint velocity \cite{costanzi2023:personal_myo}. When paired with an impedance controller, it instead determines the rate of change of the virtual spring's equilibrium position, allowing for a more compliant interaction \cite{costanzi2021:phd}.

\subsection{Admittance \Modulenames}\label{sec:adapters:adm}

An admittance \modulename\ introduces a virtual dynamic system between the decoded signal and the controller reference. Typically, $d(t)$ is interpreted as a force applied to a virtual mass–damper system, whose output is used as a velocity reference \cite{lobo-prat2016:emg_impl,liu_muscle_2021,toro_ossaba2024,Kilic2017EMGBased,Bos2020:symbihand_dmd,costanzi2021:phd}:
\begin{equation}
    M_v\ddot q_{ref}(t) + B_v \dot q_{ref}(t) = d(t)
\end{equation}
where $M_v$ and $B_v$ are the virtual mass and damper respectively. 

Admittance shaping provides a physically interpretable force-to-velocity mapping and therefore allows explicit control over the dynamic response of the assistance.  However, its parameters are often selected empirically, through pilot trials, user preferences, or heuristic tuning, and different choices can substantially affect responsiveness, comfort, and interaction behavior \cite{Bos2020:symbihand_dmd, toro_ossaba2024}. As with other velocity mappings, the behavior is substantially unaffected by external perturbations, such as changing payloads. Importantly, admittance shaping should not be confused with admittance control: the former maps decoded intention into a velocity reference, whereas the latter determines the robot's motion response based on measured interaction forces.

\subsection{Adaptive Gravity Compensation (AGC) \Modulenames}\label{sec:adapters:AGC}

Adaptive gravity compensation \modulenames\ do not use the value of $d(t)$ to directly control a physical variable, but instead to estimate the payload lifted by the user and adapt the controller reference accordingly \cite{dimo2026:partial_agc,Dimo2026MyographyGravityCompensation,Nasiri2022}. A general formulation is:
\begin{align}
    \hat{m}(t) &= \Phi\!\left(d(t),{q}(t),\dot{{q}}(t),\ddot{{q}}(t)\right), \\
    c_{ref}(t) &= \hat{m}(t)\,\boldsymbol{G}\!\left({q}(t)\right)
\end{align}
where $\hat{m}(t)$ is the estimated payload mass, $\Phi(\cdot)$ denotes the estimation function, and $\boldsymbol{G}({q}(t))$ is the posture-dependent factor that maps the estimated mass into the corresponding gravity-compensation reference. Kinematic and dynamic variables may be included to account for joint configuration and dynamic effects during non-static movements \cite{Dimo2026MyographyGravityCompensation, dimo2026:partial_agc}. 

This strategy is particularly suited for lifting and load-carrying tasks in which gravity is the dominant contribution to the required effort. Unlike integral-force shaping, AGC can explicitly account for posture-dependent changes in gravitational torque. Experimental studies in healthy subjects also suggest that full gravity compensation is not necessarily the optimal choice in all conditions, as in some cases partial gravity compensation has been associated with greater comfort \cite{dimo2026:partial_agc}.

\subsection{Learning-based \Modulenames}

Learning-based approaches can represent nonlinear and time-varying transformations that depend on the user, the task, and the interaction dynamics. Recent RL frameworks, for example, use latent representations to account for factors such as external payloads, limb configuration, or subject-specific variability, and can adapt the assistive policy online \cite{karimi2025ilead,karimi2026zeroshot}. In such systems, the shaping law is not prescribed analytically but emerges from a learned policy mapping physiological and kinematic observations into an assistive command.

These approaches have the advantage of representing complex assistance policies without explicitly specifying an analytical shaping law. In the studies considered here, however, experimental validation within MCSs remains more limited than for analytical shaping strategies, and the resulting assistance policy is generally less directly interpretable in physical terms or explicitly tunable. Additionally, learning-based architectures may merge the decoder and the \modulename\ at the implementation level \cite{sedighi2025meta}. Nevertheless, this does not eliminate their functional distinction: the portion of the learned mapping that determines how the decoded intention is transformed into an assistive action retains the functional role of the \modulename.

\begin{table*}[ht]
\centering
\small
\setlength{\tabcolsep}{4pt}
\renewcommand{\arraystretch}{1.25}

\rowcolors{3}{gray!10}{white}

\caption{
Comparative summary of \modulename\ types in terms of achievable assistance level, strengths, and limitations.
}

\begin{tabularx}{\textwidth}{
    L{2.1cm}
    L{3.2cm}
    X
    X
}
\toprule
\rowcolor{white}
\textbf{\Modulename\ Type} &
\textbf{Assistance Level $\eta$} &
\textbf{Strengths} &
\textbf{Limitations} \\
\midrule

\textbf{Proportional} (force$\to$force) &
$\eta \approx \frac{\beta}{1+\beta}$ &
Simple and interpretable; $\eta$ explicitly computable; widely validated on humans & Instability/performance degradation reported above $\eta \approx 0.5-0.6$ \cite{Kwon2014,missiroli2020:assistance_level}\\

\textbf{Proportional} (force$\to$velocity) &
Not directly computable; variable, $\eta = 1$ in static conditions, decreases with velocity &
Payload-independent behavior &
No established $\eta$ guidelines; no established calibration guidelines \\

\textbf{Integral} (force$\to$force) &
$\eta \to 1$ achievable for quasi-static tasks &
Reduces effort in static conditions &
Assistance level not adjustable; not suitable for fast tasks; not robust to arm kinematics, since no gravity-compensation law is applied \\


\textbf{Admittance} &
Not directly computable &
Physiologically motivated force$\to$velocity mapping; particularly suitable for users with no residual movement, e.g., pathological populations \cite{lobo-prat2016:emg_impl} &
Requires careful tuning of virtual dynamics, often selected empirically or according to user preferences; inappropriate parameters may affect responsiveness, stability, comfort, and interaction compliance \\

\textbf{Adaptive gravity compensation} &
$\eta \to 1$ achievable for quasi-static tasks; partial gravity compensation possible &
Robust to arm kinematics through the gravity-compensation model; robust to subject anthropometry; relatively robust performance under dynamic conditions &
Partial gravity compensation has been shown to be preferred over full compensation \cite{ramella2024:passive_gc_level_eval,dimo2026:fob}; optimal partial compensation level has not yet been identified \\

\textbf{Learning-based} &
Not directly computable; output of the learning algorithm &
Can encode complex nonlinear mappings; can optimize assistance policies &
$\eta$ not directly tunable; \modulename/decoder boundary may collapse \\

\bottomrule
\end{tabularx}

\label{tab:shaper_comparison}
\end{table*}

\subsection{Discussion On \Modulenames}\label{sec:adapters:discussion}

The shaping strategies discussed above can be viewed as different implementations of a general mapping between the decoded intention $d(t)$ and the controller reference $c_{\mathrm{ref}}(t)$. This mapping may depend not only on the current decoder output, but also on its past values and additional information, such as joint kinematics or estimated task variables.

A central function of the \modulename\ module is to regulate the level of assistance. Because assistance level is not defined consistently across the literature, we use the following force-sharing index as an operational measure:
\begin{equation}
  \eta = \frac{F_{exo}}{F_{h} + F_{exo}}
  \label{assistance_definition}
\end{equation}
where $F_{exo}$ and $F_{h}$ are the assistive and the human-generated force, respectively. In the case of a \emph{proportional} force-to-force \modulename, where $d(t)$ is calibrated to track human force and scaled by a gain $\beta$, the nominal assistance level can be explicitly computed as:
\begin{equation}
  \eta = \frac{F_{exo}}{F_{h} + F_{exo}} = \frac{\beta d(t)}{F_{h}+\beta d(t)} \simeq  \frac{\beta}{1+\beta}.
  \label{assistance_definition_prop}
\end{equation}
This result is valid if in the considered domain the decoded signal $d(t)$ remains a good estimate of $F_{h}$. For other shaping laws, such as force-to-velocity or velocity-to-velocity mappings, a direct relationship between $\eta$ and $\beta$ cannot be computed. In these cases, the resulting force-sharing depends on the combined human–exoskeleton dynamics and the task: the assistance level must therefore be assessed experimentally, for example by measuring human and robot forces and applying \eqref{assistance_definition}.

Although several studies suggest that an optimal assistance level may exist under specific conditions \cite{missiroli2020:assistance_level}, identifying and achieving it remains an open challenge. Due to the strong coupling between the human sensorimotor system and the exoskeleton, the appropriate assistance level depends on individual and task-related factors, including physiology, motor coordination, fatigue, perceived comfort, device acceptance, and user adaptation.
Moreover, in the literature the level of assistance is rarely explicitly quantified or reported in a way that allows for direct comparison between different MCSs (Table~\ref{tab:mcs_classification}). 

Experimental studies have reported performance degradation and stability issues at high assistance levels in proportional force-to-force control. In particular, increasing the assistive torque to levels approaching or exceeding the human-generated torque may destabilize the coupled human–robot system \cite{Kwon2014}, suggesting a stability limit around $\eta \approx 0.5$ under the investigated conditions. However, other studies have reported movement smoothness comparable to unassisted conditions at a nominal force-sharing ratio of $\eta \simeq 0.55$ (estimated here), although the system configuration differed and higher assistance levels were not investigated due to hardware limitations \cite{missiroli2020:assistance_level}.

By contrast, adaptive gravity compensation strategies have demonstrated stability even at $\eta \approx 1$ in slow or quasi-static tasks \cite{Nasiri2022,Dimo2026MyographyGravityCompensation,dimo2026:partial_agc}. 
These observations suggest that assistance limits should not be interpreted as universal properties of the exoskeleton but as properties of the coupled human-robot system and, potentially, of the chosen shaping law.

From a control-theoretic perspective, proportional force-to-force shaping can be interpreted as increasing the loop gain of the coupled human–robot sensorimotor system (Figure~\ref{fig:taxonomy:human_ctrl_exo}). As is well established in control theory, increasing loop gain in the presence of delays can reduce stability margins. Physiological delays, such as EMD, together with system delays, may therefore help explain the stability issues observed at high assistance levels. However, whether this mechanism quantitatively accounts for the experimentally reported assistance limits remains to be established.

Current approaches for designing and tuning the \modulename\ module remain largely empirical \cite{missiroli2020:assistance_level,treussart2019:controlling,lotti2020:myocontrol_soft}. This contrasts with the decoder and controller modules, which can rely on established methodologies based on fitting and control theory. In learning-based approaches, the shaping law emerges from optimization of task- and interaction-dependent objectives \cite{karimi2025ilead,karimi2026zeroshot}. Assistance
modulation becomes implicitly embedded in the learned behavior, but also strongly dependent on subject, task, and operating conditions of the training dataset.

Designing and tuning the \modulename\ is difficult because it directly and critically influences the human sensorimotor loop. The central open challenge is therefore how to design the \modulename\ as a human-in-the-loop assistance policy rather than as an empirically tuned signal transformation. Human-in-the-loop optimization can be performed in vivo, but this is time-consuming and affected by substantial inter- and intra-subject variability \cite{Diaz2023_HILO}. In silico optimization could reduce experimental burden, but requires reliable predictive models of human sensorimotor adaptation that remain difficult to model, personalize, and validate \cite{Denayer2025}, and the mechanisms underlying human adaptation to assistive devices are not yet fully understood \cite{Poggensee2021_exo_adaptation}.


\section{The Controller Module}\label{sec:controllers}

The controller module is responsible for realizing the reference $c_{ref}(t)$ generated by the \modulename\ through low-level control of the actuators. Common approaches include torque/force, velocity, position, impedance or admittance control. These strategies are well established in robotics and have already been extensively reviewed \cite{calanca2016:review_compliant_control}.
Within the proposed taxonomy, the controller role is the physical realization of the assistive command that has been determined by the assistance policy implemented by the \modulename.

\subsection{Torque Controllers}\label{sec:controllers:force_ctrl}

Torque control is widely used in assistive devices and can be implemented either implicitly, through actuator current or equivalent feedforward relationships, or explicitly, through feedback of the interaction torque or force. Implicit torque control can be effective when transmission dynamics, friction, and reflected inertia are sufficiently small and the actuator is highly backdrivable. Explicit torque control instead employs force or torque sensing or estimation to compensate for robot, actuator, and transmission dynamics, such as inertia and friction, thereby improving interaction transparency.

Mechanical compliance can substantially affect force controllability, interaction stability, and robustness to uncertainties in the human-robot-environment dynamics and should therefore be carefully considered jointly with the control design \cite{calanca2022enhancing, Calanca2016:on_the_role}.
Series Elastic Actuators (SEAs), which place an elastic element in series with the actuator \cite{Pratt1995:SEA}, are therefore particularly relevant to wearable robotics and are still widely adopted in exoskeleton designs \cite{Rinaldi2026:flexo,pan2022nesm,huo2020force,lin2020decoupled}. 

Explicit force control can achieve zero steady-state tracking error without requiring integral action, as a unit feedforward term combined with PD feedback is sufficient \cite{Dimo2026_forcecast}. Although integral action can enhance disturbance rejection, it may also reduce passivity margins \cite{Vallery2007:SEA}. In force-controlled wearable robots, friction represents a major source of disturbance, motivating the development of alternative compensation strategies. Recent studies have investigated friction compensation methods specifically designed to preserve passivity and interaction stability \cite{Dimo2024:ea_fob,dimo2026:fob}.

Despite the extensive literature on interaction control, low-level torque loops in myoelectric-driven upper-limb exoskeletons are generally implemented through relatively conventional strategies, such as current control or classical PD/PID-like feedback (see Table \ref{tab:mcs_classification}). More advanced torque-control approaches have received limited attention in the exoskeleton literature, mostly in relation to specific actuation or transmission technologies. Examples include disturbance-observer-based and adaptive strategies to improve torque tracking under uncertain and time-varying dynamics \cite{huo2020force,lin2020decoupled}, as well as sliding-mode strategies for cable-tension regulation in cable-driven devices\cite{li2023improved}. Nevertheless, these advanced solutions remain comparatively uncommon in complete MCS architectures.

\subsection{Motion, Impedance and Admittance Controllers}\label{sec:controllers:impedance_admittance}

Position and velocity controllers are conventional approaches that enforce predefined trajectories with high apparent stiffness, which can be undesirable in pHRI. In this context, where safety, comfort, and adaptability are required, compliant behavior is preferred.
Impedance and admittance controllers are commonly adopted strategies to introduce compliance. Both approaches regulate the dynamic relationship between force and velocity, allowing the user to deviate from the reference trajectory in a controlled and predictable manner \cite{hogan1985:impedance_theory}. 
They can be interpreted as rendering a virtual mechanical behavior characterized by tunable stiffness, damping, and, when relevant, inertia parameters. The main difference lies in the direction of the controlled causal relationship. Impedance control maps motion deviations into interaction forces or torques, whereas admittance control maps measured interaction forces or torques into motion references.

In exoskeletons, controllers are commonly implemented through standard position/velocity PD/PID loops or conventional impedance/admittance schemes (see Table \ref{tab:mcs_classification}). More advanced motion-control strategies remain an active research topic. Examples include sliding-mode \cite{fallaha2021sliding}, disturbance-observer-based \cite{pont2022robust}, adaptive admittance \cite{wu2019adaptive}, passivity-based \cite{zou2022robust}, and variable-impedance strategies \cite{Wang2026VariableImpedance}.

Finally, we identified a common terminology issue relevant when classifying MCSs: some velocity-sourced torque controllers have been described as admittance controllers \cite{missiroli2020:assistance_level,lotti2022:myo_vs_force}. However, if the velocity loop merely serves to regulate torque rather than to explicitly implement a true admittance behavior, the architecture should not be classified as admittance control.

\subsection{Gravity Compensation}

Gravity compensation (GC) is best regarded as a torque contribution rather than a control strategy per se. It can therefore be incorporated into torque/force, impedance, or admittance controllers  as a feedforward component of the reference. By directly compensating for gravitational loads, which are the dominant component the wearer has to overcome, GC can provide intuitive assistance without relying on sEMG signals that can be affected by noise and artifacts \cite{costanzi2023:personal_myo}.
GC may compensate for the weight of the user’s limb segments, the exoskeleton structure, or an external payload. Fixed model-based compensation can be effective when the relevant masses are known, for example, from anthropometric models and user-specific measurements \cite{chiaradia2018:soft_exo, lotti2022:myo_vs_force}. However, unknown or changing payloads require online estimation or adaptation using, for example, sEMG signals \cite{Nasiri2022,Dimo2026MyographyGravityCompensation}.

Within the proposed taxonomy, the functional role of GC depends on how its reference is generated. A prescribed gravitational torque based on a fixed model constitutes a component of the controller \cite{buongiorno_linear_2018, Sedighi2024,sedighi2025meta}. Conversely, when the gravitational contribution is estimated or adapted online based on the user’s behavior, the estimation process belongs to the \modulename\ module because it determines the assistance command (see Section \ref{sec:adapters:AGC}), while the controller is solely responsible for tracking the resulting torque reference \cite{Dimo2026MyographyGravityCompensation}.

\begin{sidewaystable*}
\centering
\small
\setlength{\tabcolsep}{3.0pt}
\renewcommand{\arraystretch}{1.03}

\rowcolors{3}{gray!10}{white}

\caption{
Representative MCSs classified according to the proposed decoder-\modulename\ -controller taxonomy.
\textit{Joints} indicates the considered joints: S = shoulder, E = elbow, W = wrist, Hd = hand.
\textit{Ass. Level} reports the assistance level when it is explicitly available or, if marked with $^*$, it can be estimated from the data provided in the corresponding paper; otherwise, it is marked as N/A.
\textit{Application} indicates the target use case: R = rehabilitation, A = assistance.
\textit{Validation} distinguishes experiments with human participants (Exp), offline validation on real human data (Off), and simulation-only studies (Sim).
\textit{Population} indicates whether the validation involved healthy (Hea) or impaired (Imp) participants.
\textit{n} denotes the number of participants or patients included in the validation.
}

\begin{tabular}{
    p{0.95cm}
    p{0.9cm}
    p{1.4cm}
    p{2.1cm}
    p{2.75cm}
    p{3.0cm}
    p{0.95cm}
    p{1.4cm}
    p{2.1cm}
    c
    c
    c
    c
}
\toprule
\rowcolor{white}
\textbf{Ref.} &
\textbf{Year} &
\textbf{Decoder} &
\textbf{Fitting Target} &
\textbf{\Modulename} &
\textbf{Controller} &
\textbf{\#\dof} &
\textbf{Joints} &
\textbf{Ass. Level ($\eta$)} &
\textbf{Application} &
\textbf{Validation} &
\textbf{Population} &
\textbf{\textit{n}} \\
\midrule

\redcite{lobo-prat2016:emg_impl}
& 2016 & Empirical & -- & Admittance & Position (PD)
& 1 & E & N/A & A & Exp & Imp & 3 \\

\redcite{Kilic2017EMGBased}
& 2017 & ML & Force & Admittance & Velocity (PI)
& 1 & W & $\approx$ $1^*$ & A & Exp & Hea & 1 \\

\redcite{Lambelet2017_eWrist}
& 2017 & Empirical & -- & Admittance & Velocity (PD)
& 1 & W & N/A & R & Exp & Hea & 1 \\

\redcite{buongiorno_linear_2018}
& 2018 & NMS & Joint torque & Admittance & Position (PD) + GC
& 2 & S; E & N/A & R; A & Off & Hea & 7 \\

\redcite{lu2019:emg_torque}
& 2019 & NLR & Joint torque & Proportional & Velocity (PID)
& 1 & E & N/A & A & Exp & Hea & 5 \\

\redcite{wu_neural_network_enhanced_2019}
& 2019 & NLR & Joint torque & Proportional & Torque (P) + Velocity (PID)
& 1 & E & $\approx$ $0.39^*$ & A; R & Exp & Hea & 5 \\

\redcite{xiao2019:prop_cable_exo}
& 2019 & ML & Joint angle & -- & Position (PID) + FF
& 1 & E & N/A & A & Exp & Hea & 9 \\

\redcite{missiroli2020:assistance_level}
& 2020 & NMS & Joint torque & Proportional & Torque (PID)
& 1 & E & $\approx$ $0.38^*$; $0.47^*$; $0.55^*$ & A & Exp & Hea & 10 \\

\redcite{lotti2020:myocontrol_soft}
& 2020 & NMS & Joint torque & Proportional & Torque (PID)
& 1 & E & N/A & A & Exp & Hea & 6 \\

\redcite{Bos2020:symbihand_dmd}
& 2020 & Empirical & -- & Admittance & Position (PID)
& 1 & Hd & N/A & A & Exp & Imp & 1 \\

\redcite{Lotti2020_intention_detection}
& 2020 & NMS & Joint torque & Proportional & Torque (PID)
& 1 & E & $0.375^*$ & A & Exp & Hea & 4 \\

\redcite{TreussartVignais2020}
& 2020 & NLR & Joint torque & Integral & (Implicit) Torque 
& 1 & E & $\approx$ $1^*$ & A & Exp & Hea & 10 \\


\redcite{Yang2021_bilateral_rehab}
& 2021 & ML & Joint angle & -- & Position (PID)
& 1 & E & N/A & R & Exp & Hea & 10 \\

\redcite{lotti2022:myo_vs_force}
& 2022 & NMS & Joint torque & Proportional & Torque (PID)
& 1 & E & N/A & A & Exp & Hea & 8 \\

\redcite{oghogho2022deep}
& 2022 & Empirical & -- & RL-policy (adaptive prop. gain) & (Implicit) Torque
& 3 & S; E & N/A & A & Exp & Hea & 1 \\

\redcite{Nasiri2022}
& 2022 & NMS & Joint torque & Adaptive GC & (Ideal) Torque
& 2 & S; E; & $\approx$ $1^*$ & A & Sim & -- & -- \\

\redcite{Sierotowicz2022_hand}
& 2022 & ML & Grasping force & Admittance & Velocity (PID)
& 2 & Hd & N/A & R; A & Exp & Hea & 6 \\

\redcite{Li2023SubjectIndependent}
& 2023 & ML & Joint angle & -- & Position (PID)
& 1 & E & N/A & R & Exp & Hea & 10 \\

\redcite{wu2023semg}
& 2023 & NLR & Joint torque & Adaptive admittance & Position (SMC)
& 1 & E & N/A & R & Exp & Hea; Imp & 8 \\

\redcite{Li2023_torque_prediction}
& 2023 & ML & Joint torque & Proportional & (Implicit) Torque
& 1 & E & N/A & R & Exp & Hea & 7 \\

\redcite{costanzi2023:personal_myo}
& 2023 & MVLR & Joint torque & Proportional; integral & Force (PD); position (PID); impedance
& 1 & E & $\approx$ $0.71^*$; $0.75^*$ & A & Exp & Imp & 1 \\

\redcite{Han2024SAPMFCS}
& 2024 & ML & Joint angle & -- & Position (SMC)
& 1 & E & N/A & R & Exp & Hea & 4 \\

\redcite{Sedighi2024}
& 2024 & ML & Joint angle & -- & Position (PD) + GC
& 3 & S; E & N/A & A & Exp & Hea & 2 \\

\redcite{toro_ossaba2024}
& 2024 & ML & Joint torque & Admittance & Position (MRAC)
& 1 & E & $\approx$ $0.34^*$; $0.41^*$ & R & Exp & Hea & 1 \\

\redcite{Peng2024:myomo}
& 2024 & Empirical & -- & Proportional & Velocity (type N/A)
& 1 & E & N/A & A & Exp & Hea & 36 \\

\redcite{Li2025}
& 2025 & NMS & Joint torque & Proportional; admittance & Position (PID)
& 2 & Hd & $\approx$ $0.09^*$; $0.17^*$; $0.23^*$ & A & Exp & Hea & 4 \\

\redcite{karimi2025ilead}
& 2025 & ML & -- & RL-policy & Position (PD)
& 1 & E & N/A & A & Sim & -- & -- \\

\redcite{Zhan2025Volitional}
& 2025 & ML & Joint angle & -- & Position (PID)
& 1 & E & N/A & R & Off; Exp & Hea & 10 \\

\redcite{sedighi2025meta}
& 2025 & ML & Joint angle & -- & Position (PD) + GC
& 3 & S; E & N/A & A & Exp & Hea & 8 \\

\redcite{Cao2025NeuroFuzzy}
& 2025 & ML & Joint torque & ML & Variable impedance
& 2 & E & N/A & R & Exp & Hea & 3 \\

\redcite{Shakeriaski2026}
& 2026 & ML & Joint angle & -- & Position (PID); impedance; SMC
& 1 & E & N/A & R; A & Sim; Off & Hea & 12 \\

\redcite{karimi2026zeroshot}
& 2026 & ML & RL-policy & ML & Position (PD)
& 1 & E & N/A & A & Exp & Hea & 10 \\

\redcite{Dimo2026MyographyGravityCompensation}
& 2026 & MVLR & Joint torque & Adaptive GC & Torque (PD)
& 1 & E & $\approx$ 1 & A & Exp & Hea & 15 \\

\redcite{dimo2026:partial_agc}
& 2026 & MVLR & Joint torque & Adaptive GC & Torque (PD)
& 1 & E & $\approx$ 1 & A & Exp & Hea & 20 \\

\bottomrule
\end{tabular}

\label{tab:mcs_classification}
\end{sidewaystable*}

\section{Overall Discussion and Current Limitations}
\label{sec:examples}

Table~\ref{tab:mcs_classification} summarizes representative MCSs for upper-limb exoskeletons according to the proposed decoder-\modulename-controller taxonomy. Beyond classifying individual architectures, the table highlights several trends and limitations that have emerged in the representative literature of the last decade. Several important research questions remain open, the most relevant of which are summarized in Figure~\ref{fig:question_box}.

The first trend, highlighted by Table~\ref{tab:mcs_classification}, concerns the decoder. Earlier systems predominantly relied on empirical mappings, nonlinear regression models, or NMS approaches, whereas ML decoders have become increasingly common in recent years. However, within the considered studies, ML decoders are frequently validated offline or in simulation, rather than as part of complete real-time experiments on physical devices. The table also reveals an association between decoder type and fitting target: NMS decoders are predominantly calibrated against joint torques, while ML decoders are more frequently trained to predict joint angles and subsequently coupled to position controllers, even if they could, in principle, allow greater flexibility in the choice of the fitting target. This distribution may partly reflect the greater availability of kinematic datasets compared with datasets containing interaction forces, joint torques or internal forces.


A similar evolution can be observed for the \modulename. Earlier systems generally employed simple shaping strategies, such as proportional gains or admittance mappings. More recent approaches have increasingly incorporated more sophisticated solutions, including adaptive GC, variable-impedance, and learning-based assistance policies. At the same time, the \modulename\ is no longer explicitly identifiable in recent ML approaches, because the mapping from myoelectric signals to control references is learned jointly with the decoding process. 
This does not eliminate the functional role of the \modulename; rather, it makes its contribution more difficult to isolate and compare. Maintaining this functional distinction is therefore useful even when the corresponding implementation is embedded within another module.

Despite this evolution, Table~\ref{tab:mcs_classification} also highlights a major limitation of the current literature: the assistance level is rarely reported explicitly. In many entries, this information is marked as N/A. It is important to highlight that the assistance-level values reported in the table are not always explicitly provided by the original papers. In several cases, the values marked with $^*$ were estimated within this review from the data available in the reviewed studies, using the formulation in \eqref{assistance_definition} whenever the necessary quantities, i.e., $F_h$ and $F_{exo}$, could be retrieved. This lack of explicit reporting is critical because the assistance level is one of the most important quantities for understanding and comparing MCSs. Without it, two systems with similar decoders and controllers may provide substantially different assistance while appearing equivalent from the architectural description alone.

Table \ref{tab:mcs_classification} shows that the controller module is generally based on established and relatively simple control laws, most commonly conventional PD/PID-like position, velocity, or torque loops. More advanced control strategies, such as sliding-mode control (SMC), variable-impedance, and model-reference adaptive controllers (MRAC), remain comparatively uncommon. One aspect that has received comparatively little attention is the role of passivity. This is particularly noteworthy given that passivity provides a well-established framework for analyzing the stability of pHRI in the presence of uncertainties, especially those arising from human dynamics \cite{colgate1988:contact,Hogan2022}. Importantly, preserving passivity does not necessarily require sophisticated control strategies. For example, explicit force control can be implemented using PD feedback combined with feedforward action, without integral action, while friction compensation can be designed to preserve passivity (see Section \ref{sec:controllers:force_ctrl}). The limited attention devoted to these established principles highlights an opportunity to strengthen the connection between MCSs and fundamental results from control theory.





Finally, validation protocols remain heterogeneous and are often limited in scope. Most experimental studies involve only healthy participants, while very few studies have included participants with impairments; in any case, sample sizes are generally limited.
Many systems are also evaluated on a single degree of freedom configuration, with elbow control particularly prevalent. Multi-\dof\ architectures exist, but their validation remains comparatively limited. These factors restrict the extent to which reported improvements can be generalized across users, tasks, payloads, and clinical or assistive populations.

Taken together, the representative literature supports the usefulness of the proposed functional decomposition. Recent progress has been particularly evident in the decoder and, more recently, in adaptive \modulenames. At the same time, it reveals important limitations: assistance level is rarely quantified, most systems remain single-\dof\ and elbow-centered, passivity is seldom explicitly considered in controller design, and validation on impaired users is still limited. These observations motivate the design guidelines and research directions discussed below.

\begin{figure}[htp]
\centering
\begin{tikzpicture}
    \def\boxwidth{0.95\linewidth}

    \node[
        draw=orange,
        line width=1pt,
        rounded corners=8mm,
        fill=white,
        minimum width=\boxwidth,
        minimum height=110mm,
        inner xsep=5mm,
        inner ysep=4mm,
        align=center
    ] (box) {};

    \node[
        text=orange,
        font=\bfseries
    ] at ([yshift=-5.5mm]box.north) {
        Outstanding Questions
    };

    \draw[
        orange,
        line width=0.8pt
    ] 
    ([xshift=3mm,yshift=-10mm]box.north west) --
    ([xshift=-3mm,yshift=-10mm]box.north east);

    \node[
        anchor=north west,
        align=left,
        text width=\dimexpr\boxwidth-10mm\relax,
        text=black
    ] at ([xshift=5mm,yshift=-10mm]box.north west) {
    \begin{itemize}[leftmargin=0mm]
        \item How can large-scale multimodal datasets support models that generalize across users, tasks, payloads, and sessions?
        \item Can myoelectric decoders that generalize across users, tasks, payloads, and sessions with minimal subject-specific calibration be developed?
        \item How should an assistance policy, for a given user and task, be designed and tuned online while accounting for human adaptation, task requirements, dynamics, comfort, and stability?
        \item How can the stability properties of sEMG-based human-in-the-loop control be theoretically characterized, accounting for electromechanical delay, controller dynamics, and human adaptation, and leveraged to derive design rules for shaper modules?
        \item How can human-in-the-loop optimization be translated from simulation to safe real-world deployment?
        \item Which benchmarking protocols are needed to enable reproducible comparison of complete MCS architectures?
    \end{itemize}
    };
\end{tikzpicture}
\caption{Summary of the main open research questions identified through the critical analysis of the reviewed literature.}\label{fig:question_box}
\end{figure}

\section{Design Guidelines}\label{sec:guidelines}

The proposed taxonomy not only serves as a classification tool but also provides practical principles for the design of MCSs. A main implication of our modular taxonomy is that decoder, \modulename\, and controller should be designed as distinct modules and tuned separately, so that the role of each component remains interpretable. Other guidelines derive from the critical analysis of existing works and related shortcomings. 

\subsection{Modular Tuning Strategy}


\guideline{Tune the decoder first.}
The decoder should be calibrated and validated before tuning the shaper and controller. Its performance should be evaluated in terms of fitting accuracy with respect to the selected decoding target and, where relevant, robustness and generalization across task-relevant conditions, such as joint configurations, payloads, fatigue states, and electrode placements. Calibration time and real-time computational cost should also be considered when relevant to the application.

\guideline{Tune the controller second.}
The controller should be designed and validated independently, with its performance evaluated in terms of tracking accuracy, bandwidth, stability, disturbance rejection, and phase lag over the task-relevant frequency range. Where appropriate, human-interaction effects may be incorporated as disturbances or system uncertainties and analyzed using suitable stability or passivity-based methods. 

\guideline{Tune the \modulename\ last.}
The \modulename\ should be tuned only after fixing the decoder and controller. Since it directly alters the human sensorimotor loop, its parameters should be selected through human-in-the-loop testing or, when this is not feasible, using a validated surrogate model. This allows changes in performance, comfort, stability, or effort reduction to be attributed to the assistance strategy itself.


\subsection{Decoder Guidelines}

\guideline{Prefer physically meaningful decoder outputs.}
Whenever possible, the decoder should be calibrated against a measurable biomechanical quantity representing the user’s intended action, such as joint torque, force, or velocity. Arbitrarily scaled or dimensionless outputs should be avoided unless explicitly justified, as they make decoder performance difficult to assess. A well-defined decoding target improves interpretability, enables comparison across decoders, and provides a clear interface for \modulename\ design.

\guideline{Prefer torque- or force-related targets when load-robust assistance is the objective.}
Direct sEMG-to-position decoding should be used with caution when the objective is load-robust assistance, because the relationship between muscle activation and joint position is strongly task- and load-dependent. For assistive applications in which load compensation is important, force- or torque-related targets may therefore provide a more physically meaningful interface.
When a kinematic target is required, joint velocity should generally be preferred over position, or an admittance \modulename\ should be used (see Section \ref{sec:adapters:adm}).

\guideline{Match decoder complexity to the application.}
Simple regression models should be considered as baseline approaches, particularly when calibration time is limited. NMS decoders may offer advantages when physiological interpretability or consistency across multiple degrees of freedom is a priority. ML decoders provide greater flexibility in learning complex mappings, but their adoption requires careful evaluation of training data coverage and the ability to generalize to unseen tasks, payloads, and users. To the best of the authors' knowledge, robust generalization of ML-based decoders across users, tasks, payloads, and experimental sessions remains insufficiently demonstrated in existing MCS studies.

\subsection{\Modulename\ Guidelines}

\guideline{Keep the \modulename\ explicit whenever possible.}
The shaping function should remain clearly identifiable, even when integrated into the decoder or controller. Its functional role remains distinct, even when implemented as a simple gain, as it separates assistance shaping from intention decoding and actuator control, thereby facilitating the comparison of different assistance strategies.

\guideline{Report the assistance level and \modulename\ tuning criteria.}
The assistance level should be explicitly reported whenever it can be estimated. When direct computation is not possible, the study should at least report the \modulename\ parameters and measures of human and robot contributions during operation. The criteria used to tune the \modulename's parameters should be reported as well. This should become standard practice.


\subsection{Controller Guidelines}

\guideline{Match controller compliance to the interaction and therapy mode.}
High-stiffness position control may be unsuitable for assistive pHRI applications unless sufficient compliance is ensured through mechanical design or control strategies. Impedance, admittance, or torque control may instead be preferable, as they allow users to deviate from the reference trajectory in a controlled manner while promoting safer interaction. Nevertheless, position control may remain appropriate in passive neurorehabilitation, particularly when patients are unable to actively contribute to the movement and predefined trajectories must be reproduced.

\guideline{Minimize unnecessary controller delay.}
The controller should be designed to minimize phase lag within the frequency range relevant to voluntary upper-limb movements. This limits the delay introduced by the low-level control loop, making the remaining EMD and computational delays easier to characterize and compensate for.

\guideline{Match the force control strategy to actuator mechanics.}
The choice between implicit and explicit force control should account for actuator inertia, friction, transmission dynamics, and backdrivability. Implicit force control may be effective when inertia and friction are sufficiently low, whereas explicit force control may be more suitable when precise force regulation or high interaction transparency is required. Regardless of the approach, low actuator inertia and friction generally improve force controllability \cite{calanca2018:acc_feedback, calanca2022enhancing}.


\section{Future Directions}\label{sec:future_directions}

The proposed taxonomy suggests that future progress in MCSs will depend not only on improved decoding algorithms, but on a better integration of biomechanical modeling, assistance shaping, interaction control, and validation protocols. Four research directions appear particularly important.

\futuredir{Dynamics-enriched datasets}
Large-scale multimodal and dynamics-enriched datasets are needed to move beyond predominantly kinematic EMG decoding in ML approaches. Robotic assistance is not equivalent to motion prediction: the same trajectory may require substantially different joint torques depending on payload, weakness, compensatory strategies, and anthropometry. Future datasets should therefore combine synchronized sEMG, kinematics, interaction forces, joint torques, contact information, payload conditions, and subject-specific biomechanical descriptors. Low-cost, markerless MOCAP systems such as OpenCap may facilitate scalable access to a large amount of data acquired outside traditional laboratories and support the development of torque-informed datasets \cite{Uhlrich2023OpenCap,Turner2024OpenCap}.

\futuredir{Transferable and biomechanics-informed pretrained models}
Large and diverse datasets could support the development of pretrained models that generalize across subjects, tasks, payloads, and devices, thereby reducing the need for long subject-specific calibration. Rather than training each decoder from scratch, future MCSs could combine pretrained representations of muscle activity, movement intention, effort, or task demand with lightweight personalization using Transfer Learning (TL) or domain adaptation approaches \cite{wei2024:emg_review}. Existing motion foundation models such as MotionBERT demonstrate the potential of large-scale pretraining, but primarily capture geometric and kinematic relationships \cite{Zhu2023MotionBERT}. Wearable robotics requires models that also capture the relationship between movement, load, effort, and torque generation. Biomechanics-informed foundation models, possibly based on self-supervised or predictive learning paradigms\cite{Assran2023IJEPA}, therefore represent a promising research direction.

\futuredir{Stability-aware assistance shaping}
Proportional shapers (Section~\ref{sec:adapters}) are widely used because of their simplicity and interpretability, but the assistance gain cannot be increased indefinitely without degrading the stability of the coupled human-robot system. Future work should move beyond empirical gain tuning and derive design criteria that explicitly account for EMD, controller bandwidth, noise, human adaptation, and interaction dynamics. Such criteria could define safe operating regions for both the timing and the magnitude of assistance and provide a theoretical basis for the design of assistance policies.

\futuredir{Real-world human-in-the-loop optimization and benchmarking}
Human-in-the-loop optimization (HILO) and reinforcement-learning (RL) approaches are increasingly explored in simulation or offline settings, whereas the performance of an MCS ultimately depends on the coupled dynamics of user, exoskeleton, environment, and assistance policy. Future systems should therefore bring HILO to real devices under explicit safety constraints. In rehabilitation, this could enable assistance to maintain the user within a desired range of active motor participation; in occupational applications, it could adapt support to changing payloads and task conditions.
At the same time, standardized benchmarking protocols for complete MCS architectures are needed. Evaluation should extend beyond decoder accuracy to include assistance effectiveness, stability, robustness across sessions, generalization across payloads, subjects, and tasks, interaction safety, effort and fatigue, comfort, smoothness, perceived agency, and consistency between human intention and robotic action. 
Environmental interaction should also be incorporated into these protocols, particularly for contact-rich and object-manipulation tasks \cite{xiong2024:intuitive_hrei_emg}. \smallskip

Overall, the long-term objective is not simply to predict the movement a user intends to perform, but also to determine how much assistance is required, when it should be delivered, and how it can remain stable and synergistic with both the user and the environment.

\section{Conclusion}\label{sec:conclusions}

This review examined myoelectric control systems (MCSs) for upper-limb exoskeletons through a functional three-module framework comprising a decoder, a \modulename, and a controller. This decomposition separates intention estimation, assistance shaping, and low-level actuation, providing a common basis for interpreting heterogeneous architectures and deriving design guidelines.

The reviewed literature indicates substantial progress in myoelectric decoding, while assistance shaping remains comparatively less formalized and often difficult to distinguish from the decoder or controller. 
Major open challenges include the limited reporting of assistance levels, limited consideration of interaction-aware control and passivity, the prevalence of single-\dof\ and healthy-user validation, and insufficient evidence of generalization across users, tasks, and payloads.

Future progress will require dynamics-enriched datasets, pretrained models that combine data-driven learning with biomechanical structure, stability-aware assistance policies, and human-in-the-loop optimization and benchmarking on real-world devices.

\bibliographystyle{IEEEtran}
\bibliography{IEEEabrv,bibreport}

\end{document}